# Aligning Tutor Discourse Supporting Rigorous Thinking with Tutee Content Mastery for Predicting Math Achievement

Mark Abdelshiheed, Jennifer K. Jacobs, and Sidney K. D'Mello
{mark.abdelshiheed, jennifer.jacobs, sidney.dmello}@colorado.edu

University of Colorado Boulder, Boulder, CO 80309, USA

**Abstract.** This work investigates how tutoring discourse interacts with students' proximal knowledge to explain and predict students' learning outcomes. Our work is conducted in the context of high-dosage human tutoring where 9th-grade students ($N = 1080$) attended small group tutorials and individually practiced problems on an Intelligent Tutoring System (ITS). We analyzed whether tutors' talk moves and students' performance on the ITS predicted scores on math learning assessments. We trained Random Forest Classifiers (RFCs) to distinguish high and low assessment scores based on tutor talk moves, student's ITS performance metrics, and their combination. A decision tree was extracted from each RFC to yield an interpretable model. We found AUCs of 0.63 for talk moves, 0.66 for ITS, and 0.77 for their combination, suggesting interactivity among the two feature sources. Specifically, the best decision tree emerged from combining the tutor talk moves that encouraged rigorous thinking and students' ITS mastery. In essence, tutor talk that encouraged mathematical reasoning predicted achievement for students who demonstrated high mastery on the ITS, whereas tutors' revoicing of students' mathematical ideas and contributions was predictive for students with low ITS mastery. Implications for practice are discussed.

## 1 Introduction

While "silence is one great art of conversation" is a popular quote in philosophy, a greater art in education is having productive conversations, as we learn to speak so we can speak to learn. Decades of research on classroom discourse have supported similar findings across a range of domains, including math, science, and English language arts [21, 22, 25]. Indeed, rich classroom talk has been highlighted as a key component in national educational standards in math [23, 24]. Numerous theories purport that students benefit from collaborative interactions and dialogue, particularly in instructional settings [2, 7, 31]. There is emerging empirical evidence to back up these theories, suggesting that students learn more from discussion-based instruction compared to direct instruction [10, 27, 32].

However, as not every type of talk sustains learning, discourse frameworks such as transactive [19], exploratory [21], and accountable [22, 29] talk have emerged to identify and define what type of talk is most consequential. Importantly, these frameworks enable the development of computational approaches to filtering and labeling relevant utterances.

In the context of human tutoring, which is our *present focus*, high-quality discourse has been shown to support students' engagement, critical thinking, and conceptual understanding [18, 21]. However, many tutors lack training and experience in facilitating tutorials that incorporate rich discourse [30]. This work focuses on **talk moves** [16, 26], which are a key component of accountable talk. While some studies on talk moves have demonstrated their association with improved discourse and student learning [10, 27, 32], prior work has yet to directly connect, *at scale*, instructional discourse models with learning outcomes, which is one goal of the present work.

To mitigate the high costs of human tutoring, researchers and practitioners are blending human tutoring with e-learning environments, specifically Intelligent Tutoring Systems (ITSs). There is a rich history of inferring learning outcomes based on students' behaviors in these environments, including time on task and accuracy in applying problem-specific principles [8, 12] and the use of hints and problem-solving strategies [1, 5, 6, 12]. Indeed, substantial work has leveraged artificial intelligence and machine learning tools to use students' logs in predicting performance in e-learning environments [5, 14] and external assessments [8, 12, 17]. Despite the considerable success of these tools, prior research has yet to investigate how individual-based ITS performance interacts with group-based tutoring discourse in explaining and predicting students' learning outcomes.

In this work, we conduct a *large-scale evaluation* of whether the interaction of tutoring discourse (human tutor's talk moves) and students' content knowledge (performance on a math ITS) predicts students' math assessment scores. Our investigation is in the context of high-dosage, small group tutoring sessions with a human tutor, which have emerged as a key tool to address pandemic-related learning loss and help close achievement gaps [13]. Using data from 1080 9th-grade students from a large urban Midwestern district, we investigate whether an interpretable model can predict students' math achievement based on combinations of talk moves and ITS performance metrics. Our approach involves extracting a decision tree from a random forest classifier to get the benefits of high *interpretability* from the former and high *accuracy* from the latter.

### 1.1 Instructional Discourse

Productive discussions between teachers and students have long been a center of intensive study in educational research. For example, Webb et al. [32] showed that students' participation in classroom math conversations predicted their achievement. In particular, they found that teachers' encouragement of and follow-up on students' productive talk (i.e., talk moves) increased students' engagement, which cascaded to improve their learning outcomes.

Three popular perspectives to model meaningful teacher-student dialogues are transactive [19], exploratory [21], and accountable [22, 29] talk. Transactive talk involves students transforming arguments they hear by building on each other's reasoning. This process involves refuting arguments till a final, winning argument is reached based on the group discussion. Exploratory talk involves dialogue in which students offer the relevant information they have such that everyone engages critically and constructively with others' ideas, all members try to periodically reach an agreement about major ideas, and all ideas are treated as worthy of attention and consideration.

Accountable talk theory identifies and defines an explicit set of discursive techniques that can promote rich, knowledge-building discussions in classrooms [22, 29]. At the heart of accountable talk is the notion that teachers should organize discussions that promote students' equitable participation in a rigorous learning environment where their thinking is made explicit and publicly available to everyone in the classroom. Accountable talk outlines **three** general requirements of classroom discussions: accountability to the learning community, to content knowledge, and to rigorous thinking [26, 29]. In this work, we focus on ***talk moves***, which are linguistic acts that are intended to facilitate dialogue.

Prior studies have articulated that talk moves show potential in improving discourse and students' learning outcomes [10, 27, 32]. Chen et al. [10] developed a visualization tool to support teachers' reflections on classroom discourse, particularly the use of talk moves. They found that the tool significantly increased the teachers' use of productive talk moves compared to teachers who never used the tool. Additionally, students of the former set of teachers had significantly higher math achievement scores than their peers. O'Connor et al. [27] showed over two studies that teachers' use of talk moves to facilitate academically productive talk was associated with significantly higher standardized math tests in their students when compared to those who received direct instruction.

In essence, substantial research has shown the potential of rich discussions, as indicated by the use of talk moves, in improving students' learning outcomes. However, as far as we know, no attempts were made to connect group-based discourse during tutoring sessions with individual-based ITS performance.

### 1.2  Student Logs and Performance on Intelligent Tutoring Systems

Based on the considerable logistical challenges associated with offering frequent human tutoring, Intelligent Tutoring Systems (ITSs) are sought as interactive e-learning environments where students have the opportunity to individually learn in a personalized fashion [11]. Prior work has shown that tracing students' logs and overall performance on ITS is predictive of their final performance on ITSs [1, 3–6, 14, 15] and on test scores [8, 12, 17]. The traced logs include time on task and the use of problem-solving strategies, amongst others.

To predict and influence learning outcomes on ITSs, Islam et al. [15] showed that an apprenticeship learning framework effectively modeled students' pedagogical decision-making strategies on a probability ITS and impacted students' learning gains positively. Hostetter et al. [14] revealed that students with specific personality traits benefited significantly more from personalized explanations individually tailored to their pedagogical decisions. Abdelshiheed et al. [1, 3–6] found that the knowledge of how and when to use each problem-solving strategy predicted students' learning outcomes on a logic ITS and the transfer of metacognitive knowledge to a subsequent probability ITS.

In the context of predicting test scores, Baker et al. [8] found that individual student modeling frameworks performed significantly better than ensemble models on a genetics ITS for predicting students' paper test scores. Jensen et al. [17] showed that context-specific activity features extracted from interaction patterns on a math platform were significantly predictive of post-quiz performance. Feng et al. [12] tracked the students' number and average of hints received and requested to predict a standardized math assessment (i.e., the MCAS) of high school students. They found that the fitted regression models showed positive evidence of predicting the assessment scores.

In short, considerable work has leveraged students' performance on ITSs in predicting their achievement. However, as students practice individually on ITSs, it remains an open question whether their performance interacts with the instructional environment they experience when working with a human educator, such as the discourse occurring during small group tutorial sessions. In this study, we address this research question to explain and predict math assessment scores of students who receive tutoring from both an ITS and a human tutor.

### 1.3   Present Study

We investigate whether group-based human tutoring discourse interacts with the student's individual ITS performance to explain and predict the student's math achievement. We evaluate our research question in the **context** of high-dosage, small group tutorials with a human tutor on 1080 9th-grade students from a large urban Midwestern district. We focus on whether the *combination* of the tutor talk moves and the student's ITS performance can accurately predict the student's math assessment scores and changes in those scores over time.

We prioritize inducing an *interpretable* model that **accurately** explains and predicts students' math achievement in terms of the interaction between tutor talk moves and students' ITS performance. Specifically, we extract a Decision Tree (DT) from a Random Forest Classifier (RFC) to leverage the benefits of the high interpretability of a DT and the high accuracy of a RFC. We emphasize that the interpretability and accuracy of the model take higher **priority** than computational efficiency for the interest of the present work.

## 2   Methods

### 2.1   Participants

Students received tutoring during their regular school day (i.e., a second math period), where they attended small group tutorials with a human tutor every other day. Each group comprised at most six students (though most were two or three students) and was assigned to a tutor at the beginning of the year. Providing human tutors was part of a partnership between a large urban Midwestern district and *Saga Education*, a non-profit tutoring service provider. The frequency and duration of tutorials depended on many factors, such as the schedule, the tutor's pace, and the group's pace of processing information. A tutorial was typically 30 to 60 minutes long, and students received $2-3$ tutorials a week, yielding approximately $70-85$ tutorials in the school year.

Students alternated between working with the human tutor and individually working on the **MATHia** ITS from Carnegie Learning[1] to work on assigned math problems. Students practiced on the ITS without peer or human tutoring support. The ITS consists of 165 unique workspaces, each covering an algebraic or geometric topic. A workspace consists of topic-specific *skills* where a student aims to master each skill. On average, a workspace has four skills, yielding a total of 645 skills. Each workspace had a limited availability as it could be assigned only at a certain time of the year. Due to this limited availability and high content variability between workspaces, we collected the **aggregate** features of the ITS (Section 2.2). From now on, we refer to human tutors as '**tutors**' and MATHia ITS as '**ITS**' to distinguish the *human* factor from the *artificial* one.

The participants in this study are 9th-grade students who received math tutoring during the 2022-23 school year, where the majority of students ($> 80\%$) were low-income. During the school year, students completed five rounds of a math skills assessment. The assessments used or adapted items from existing measures with demonstrated reliability and validity and measured students' content knowledge at their current level as well as below-grade fundamentals. Each assessment consists of 30 questions that were graded in a binary manner, resulting in integer scores within the [0, 30] range. The five assessments occurred roughly in August, October, January, March, and May.

Students occasionally skipped assessment rounds, tutorials, and ITS sessions. Out of 1521 students, we only analyzed data from 1080 students across 46 tutors who had **at least** one assessment and attended **at least** 50% of the tutorials and 50% of the ITS sessions during the school year. Table 1 shows the distribution of completed assessments over the final set of included participants.

**Table 1.** Completed Assessment Rounds Distribution Across Students

| # Completed Assessments Per Student | # Students | # Completed Assessments |
|---|---|---|
| 1 | 112 | 112 |
| 2 | 65 | 130 |
| 3 | 127 | 381 |
| 4 | 379 | 1516 |
| 5 | 397 | 1985 |
|  | Total: **1080** | Total: **4124** |

### 2.2   Input Features: Talk Moves and ITS metrics

Tutorials were audio and video recorded as part of the standard protocol used by *Saga Education*, the tutoring service provider. To investigate how the *tutor talk* interacts with *students' proximal content knowledge*, we analyzed the tutors' talk moves from tutorials and the students' current performance on the ITS. In alignment with our goal of extracting an interpretable model, we selected a **limited** number of input features —talk moves and ITS metrics— to ensure exhausting as many possible feature combinations without computational limitations.

The talk moves framework [16, 26] categorizes an utterance made by a tutor or student into one of several labels that capture different dimensions of communication and learning. Table 2 lists the six tutor talk moves included in our study. For the scope of this work, we only focused on tutor talk moves within these categories. To automatically measure tutor talk moves, we leveraged a Robustly Optimized Bidirectional Encoder Representations from Transformers Pretraining Approach (RoBERTa)[20] model, which we fine-tuned on large data sets of talk in classrooms and a small number of Saga tutoring sessions. Details on model training and validation are discussed in Booth et al. [9]. Overall, the model was moderately accurate with a macro F1 of 0.765 for tutor talk moves.

---

[1] https://www.carnegielearning.com/solutions/math/mathia

For each student, on each completed assessment, we collected **six** features (the six tutor talk moves) for each tutorial session. For each talk move, we computed the **micro-average** *per session* during tutorials that occurred in between each assessment round (i.e., after the previous but before the current assessment date). We note that we had no way of identifying whether each tutor's talk move was directed to a specific student or a set of students in the tutorial.

**Table 2.** Description of Tutor Talk Moves

| Category | Tutor Talk Move | Description |
|---|---|---|
| Learning Community | Keeping Students Together | Orienting students to each other and to be active listeners |
| | Getting Students to Relate to Each Other's Ideas | Prompting students to react to what another student said |
| | Restating | Verbatim repetition of all or part of what a student said |
| Content Knowledge | Press for Accuracy | Eliciting mathematical contributions and vocabulary |
| Rigorous Thinking | Press for Reasoning | Eliciting explanations, evidence, thinking aloud, and ideas' connection |
| | Revoicing | Repeating what a student said while adding on or changing the wording |

Based on ITS data, we collected **five** aggregate features for each student to reflect the overall performance before each assessment round: 1) average number of mastered skills, 2) average number of opportunities needed before mastering a skill, 3) average time spent on a workspace, 4) average performance score on a workspace, and 5) average Adaptive Personalized Learning Score (APLS), which is based on a combination of the above.

### 2.3   Rationale of Decision Tree Extraction from Random Forest

Three popular, non-parametric, supervised learning algorithms are *decision trees*, *random forest classifiers*, and *random forest regressors*. A Decision Tree (DT) produces a **single**, interpretable tree from *all* features of a dataset whose response variable is *discrete*. While a Random Forest Classifier (RFC) is also trained on a *discrete* response variable, it is an **ensemble** model based on the majority vote of many DTs that were each trained on a *random subset* of the input features and data points. A Random Forest Regressor (RFR) is similar to a RFC but rather works on a *continuous* response variable.

A main advantage of DTs is their high interpretability, as the final output is a single tree that is easy to understand. However, DTs are more prone to overfitting, as they can easily overlearn the patterns to produce a perfect split of the data, resulting in models without generalizability on unseen data. In contrast, RFCs and RFRs overcome the issue of overfitting by training many trees on random subsets of features and data points. However, RFCs and RFRs lack interpretability as the majority voting mechanism is difficult to explain, especially when the number of trees comprising the ensemble model increases.

Interestingly, DTs and RFCs have an advantage —that RFRs lack— from having a discrete response variable: the **no need** for **threshold exploration**. For example, assuming a response variable of an individual's income, the discrete version will have income labeled as *high* or *low* for each individual. Therefore, a DT or RFC would simply evaluate the Gini index or information gain for each potential split on high versus low incomes. However, a continuous version of the income will have various numbers, so a RFR would have first to determine (explore) the **threshold** of distinguishing high from low incomes before training the model. The issue with threshold exploration is that it is highly prone to local minima that evaluate suboptimal rather than optimal thresholds, despite many attempts to optimize such exploration [28].

In this work, although our assessment scores are within the $[0, 30]$ range and could be treated as continuous, we avoid the RFR's threshold exploration issue by following these three steps:

1- Attempting every reasonable threshold to binarize scores into high and low.
2- Training a Random Forest Classifier (RFC) on each reasonable threshold.
3- Avoiding the RFCs' lack of interpretability by extracting a Decision Tree (DT) from each RFC, which simultaneously avoids the DT's overfitting issue.

By exploring many possible thresholds for splitting, we are **not** hypothesizing that our approach is efficient. Rather, we prioritize **exhausting** all possible combinations of extracted decision trees to find one that best explains and predicts the assessment scores from the tutor's talk moves and the tutee's ITS metrics.

### 2.4 Procedure

Figure 1a shows the data analysis mechanism for a student with no missing data. For each assessment round, the preceding six tutor talk moves and five ITS metrics features were collected as described in Section 2.2. To account for missing data (as there are only 397(37%) out of 1080 students who had data on all five assessment rounds), Figure 1b illustrates the **generic** format of collating data before each assessment round. Since assessment rounds happen on specific dates during the year, we leveraged the notion of an Evaluation Period (EP), where tutor talk moves and ITS metrics are only considered in the period that follows the previous assessment date. We had a total of 4124 EPs, one per completed assessment for each student, as suggested by the rightmost column of Table 1.

*Algorithmic Procedure:* Each EP is a record in our dataset with the input as 11 features (6 tutor talk moves + 5 student's ITS metrics) and the output as the assessment score. We extracted the decision tree using this **six-step** procedure:

1. **Binarizing Scores:** Pick a set of thresholds that distinguish high from low assessment scores. We explored thresholds from 15 to 24, as 2531(61%)

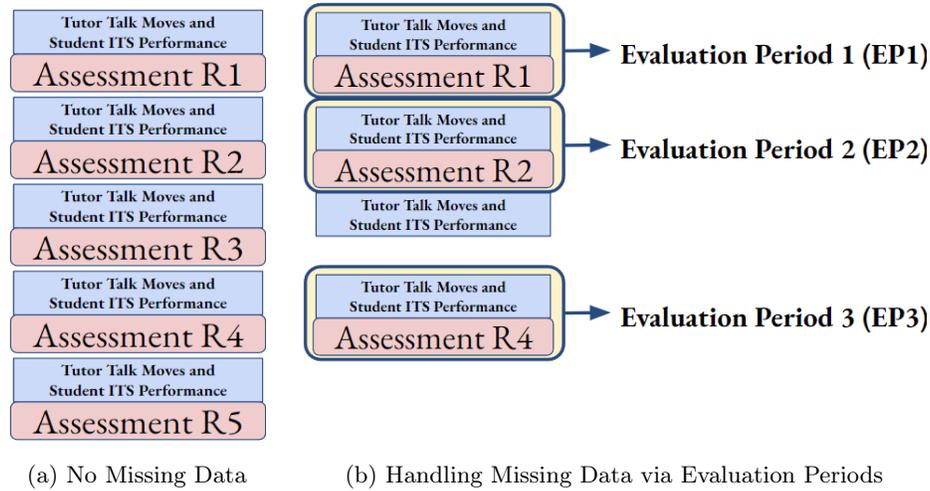

(a) No Missing Data    (b) Handling Missing Data via Evaluation Periods

**Fig. 1.** Data Collected for a Student in a School Year: Tutor Talk Moves, Student's ITS Performance, and Assessment Rounds Scores.

out of 4124 assessment scores were within that range from the original $[0, 30]$ range. Convert scores into high and low, once per candidate threshold. Values greater than or equal to the threshold become 'high,' while those below it become 'low.'

2. **Stratified, Tutor-Based[2], Five Folds for Cross Validation:** Generate 1000 random partitions of the dataset, where each partition is a five-fold candidate, and each fold has unique tutors with their students. Pick the **best** five-fold candidate that minimizes the sum of squared error of this constraint: each fold having $\approx 20\%$ of the students being tutored by $\approx 20\%$ of the tutors. The selected five-fold partition of the 46 tutors (T) and 1080 students (S) was as follows: {(T:8, S:184), (T:8, S:194), (T:10, S:225), (T:10, S:238), (T:10, S:239)}.

3. **A 60-20-20 Nested Cross-Validation:** Each fold from Step 2 acts as the **test** set *exactly once*. For the remaining four folds, one is randomly chosen as the **validation** set, and the other three as the **training** set.

4. **Training Random Forest Classifiers (RFCs):** For each binarization of scores, use the **training** set to induce $10,000$ RFCs, each with a different random seed. The hyperparameter for the number of internally generated trees was set at 10. The Gini index was used for judging the splits' quality and was computed as $G = 1 - \sum_{i=1}^{2} p_i^2$, where $p_1$ and $p_2$ are the respective probabilities of high and low labels in a given branch. The lower the Gini index, the better the split is.

5. **Extracting Decision Trees (DTs):** From each RFC, extract the DT whose decisions *agree most* with the majority vote within the RFC. Evaluate the $100,000$ extracted DTs (10 binarization thresholds X $10,000$ RFCs) on the

---

[2]Students belonging to the same tutor are in training or testing sets, but not both.

   **validation** set, and pick the DT with the **highest** AUC to represent the current fold. Repeat Steps 4 and 5 on all folds to yield *five* DTs.
6. **Final Model and Reporting Results:** Evaluate the five DTs on their **test** sets and save the test-set AUC per fold. Report the **average** 5-fold, test-set, AUC in Table 3 (rightmost column), but choose the DT from the fold with the **highest** AUC as the *final* model to evaluate on the whole dataset (Figures 2 and 3). To aid visualization, convert the labels (high vs. low) back to original numeric values and show the mean(SD) of assessment scores for each branch.

To compare our approach to standard classifiers, we appropriately modified some steps of our procedure to yield the DTs, RFCs, and RFRs shown in the middle columns of Table 3. Importantly, to induce RFRs, we skipped score binarization (Step 1), then merged and modified Steps $4-6$ to train $10,000$ RFRs based on the squared-error criterion, evaluate each on the validation set, and report the **average** 5-fold, test-set, performance. Since a RFR's default output is a coefficient of determination ($R^2$) between predictions and ground truth, we converted the $R^2$ to AUC via an effect size converter[3].

## 3 Results

We investigated *three* analyses via our algorithmic procedure: 1) how tutor talk moves and student's ITS metrics *interact* to predict assessment scores for each round, 2) whether the algorithmic procedure can accurately predict *changes* in assessment scores across rounds, and 3) how our decision tree extraction from a random forest classifier **compares** to other classifiers. Table 3 summarizes the average 5-fold AUCs of different predictors (columns) on different inputs (rows).

**Table 3.** Average 5-fold, test-set, AUCs of Talk Moves, ITS, and Their Combination

|  | Decision Tree (DT) | Random Forest Classifier (RFC) | Random Forest Regressor (RFR) | Extracted DT from RFC |
|---|---|---|---|---|
| Predicting Assessment Scores | | | | |
| Tutor Talk Moves | 0.54 | 0.58 | 0.53 | 0.63 |
| MATHia ITS | 0.56 | 0.59 | 0.54 | 0.66 |
| Combined (Talk Moves + ITS) | 0.59 | 0.63 | 0.57 | **0.77** |
| Predicting the *Changes* in Assessment Scores | | | | |
| Tutor Talk Moves | 0.55 | 0.57 | 0.55 | 0.62 |
| MATHia ITS | 0.56 | 0.59 | 0.56 | 0.64 |
| Combined (Talk Moves + ITS) | 0.57 | 0.64 | 0.58 | **0.76** |

Unlike average 5-fold AUCs here, Figures 2 and 3 evaluate final models on the *whole* dataset.

### 3.1 Interaction of Talk Moves and Intelligent Tutoring Metrics

Figure 2 shows the best-extracted decision trees from Table 3 (rightmost column, top half) but on the **whole** dataset (and hence the figures have ***higher AUCs***[4]). Coincidentally, the three trees were generated by picking the threshold of 20 for splitting high and low scores. Each root node starts from 4124 assessment rounds, and the rounds are split as we progress downwards. The Mean (SD) is shown for

---

[3] https://www.escal.site
[4] Final models yield respective AUCs of 0.64, 0.68, and 0.79 on the *whole* dataset.

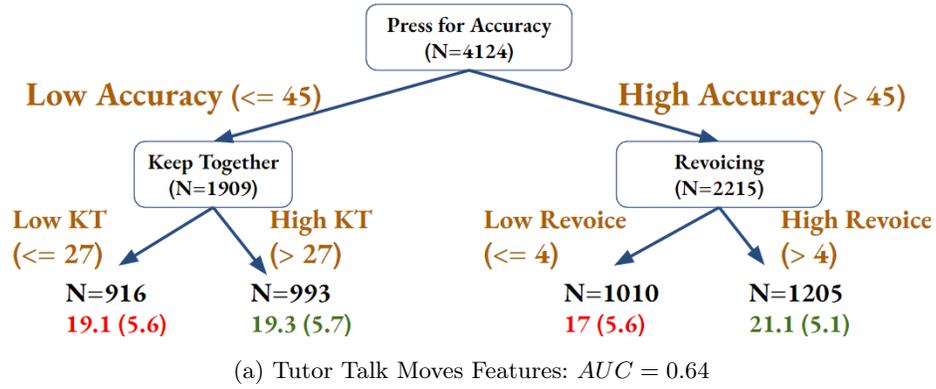

(a) Tutor Talk Moves Features: $AUC = 0.64$

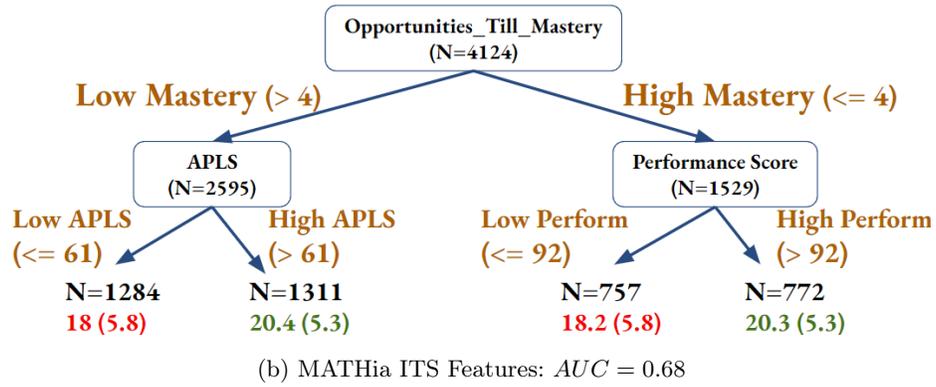

(b) MATHia ITS Features: $AUC = 0.68$

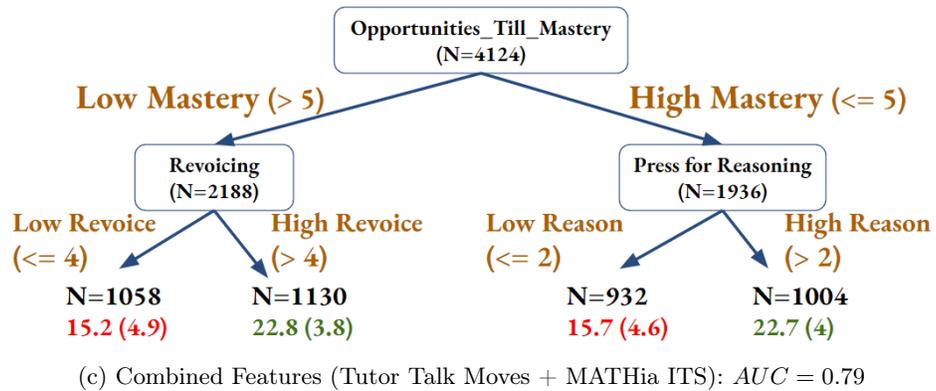

(c) Combined Features (Tutor Talk Moves + MATHia ITS): $AUC = 0.79$

**Fig. 2.** Extracted Decision Trees for Explaining Assessment Scores on **Whole Dataset**

high (in green) and low (in red) scores following the binary labeling. The AUC score ranges from 0 to 1 and reflects how well a decision tree separates high scores from low ones, where AUC of 0.5 denotes prediction by chance (luck).

Figure 2a shows that the decision tree resulting from tutor talk moves only ($AUC = 0.64$) is not very informative, despite the potentially promising right half of the tree. In particular, Press for Accuracy is the most discriminating talk move. The right half suggests that the use of Revoicing results in improved discrimination of 24% (from 17 to 21.1) for the High Press for Accuracy group. However, the left half of Figure 2a shows no score discrimination for the Keep Together talk move among the Low Press for Accuracy group.

Figure 2b illustrates that the tree based on ITS metrics only ($AUC = 0.68$) is slightly better than Figure 2a, as both halves of the tree are more stable. Specifically, for students who need more than four opportunities to master a skill —denoting low mastery— the assessment scores improve on average by 13% (from 18 to 20.4) when they have higher Adaptive Personalized Learning Score (APLS) scores. A similar 12% average improvement (from 18.2 to 20.3) occurs for high-mastery students when achieving higher performance scores.

The **best** decision tree ($AUC = 0.79$) emerges from combining all features as shown in Figure 2c. The left half of the tree shows that Revoicing significantly distinguishes scores for low-mastery students. On the other hand, the right half suggests Press for Reasoning is a significant discriminator of scores —within one to two standard deviations— for high-mastery students. In brief, on average, the assessment scores witnessed a 50% improvement (from 15.2 to 22.8) for the high usage of Revoicing for low-mastery students and a 45% improvement (from 15.7 to 22.7) for the high usage of Press for Reasoning for high-mastery students.

### 3.2 Predicting the *Changes* in Assessment Scores

To assess whether our results would persist after including students' prior assessment scores, we repeated the algorithmic procedure (Section 2.4) using the *difference* between consecutive assessment rounds as the output. Accordingly, we excluded the 112 students in the first row of Table 1 as they only had one assessment, making it impossible to compute their score change. We adjusted the candidate thresholds of splitting high versus low *changes* in assessment scores to be $[2, 2.5, 3, 3.5]$, as 68% of the changes in scores were in that range.

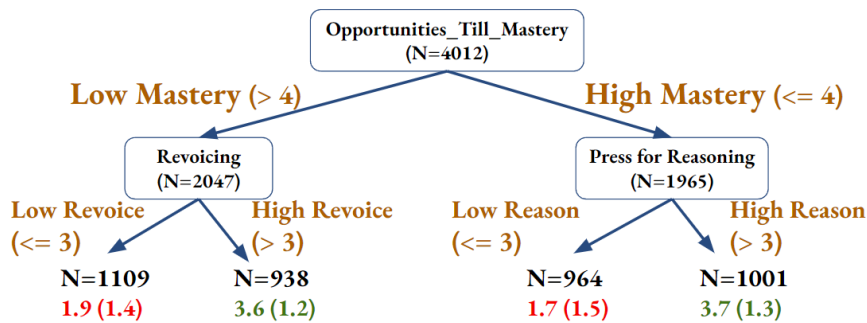

**Fig. 3.** Extracted Decision Tree for Predicting *Changes* in Scores ($AUC = 0.77$)

Figure 3 depicts the best-extracted decision tree from Table 3 (rightmost column, bottom half) but on the **whole** dataset. The tree comprised 968 students with 4012 assessments and was generated from a threshold of 2.5 for splitting high and low changes in scores. As the tree in Figure 3 used the same nodes as the tree in Figure 2c and had a comparable AUC of 0.77, this provides evidence of the robustness of the Revoicing and Press for Reasoning talk moves in discriminating assessment scores for low- and high-mastery students, respectively.

### 3.3  Comparison to Other Classifiers

To evaluate the accuracy of our procedure of extracting Decision Trees (DTs), we compared it against a traditional DT, a Random Forest Classifier (RFC), and a Random Forest Regressor (RFR), as shown in the middle columns of Table 3. None of these attempts came close to the results from the last column of Table 3. Specifically, the best traditional DT had an AUC of 0.59 and likely overfitted the data by picking suboptimal features, such as Press for Accuracy, at the root node due to its low Gini index score. The best RFC had an AUC of 0.64 and may have suffered from the majority vote mechanism, which minimized the role of meaningful trees' votes as they were a minority. Finally, the best RFR had an AUC of 0.58 (coefficient of determination: $R^2 = .018$), presumably due to the local minima resulting from the lack of picking the optimal threshold for splitting. This analysis suggests that our approach to generating a predictive model yields both robust and meaningful results.

## 4  Discussion & Conclusion

We investigated different methods of *interpreting* how a tutor's talk moves *interacts* with the student's content knowledge on a math Intelligent Tutoring System (ITS) for predicting achievement on external math assessments. Our investigation occurred in the context of high-dosage, small group human tutoring sessions that were blended with individual ITS performance. We found that extracting a decision tree from a random forest significantly outperformed the traditional classifiers in predicting students' assessment scores and changes in those scores.

The best-extracted tree emerged from combining the tutor's talk moves that encouraged rigorous thinking —Revoicing and Press for Reasoning— with the student's ITS mastery level. Specifically, tutors' use of talk moves that encouraged mathematical *reasoning* discriminated learning outcomes for students who demonstrated *high mastery* on the ITS, whereas tutors' *revoicing* of mathematical contributions and ideas was discriminating for those with lower mastery. Our findings about the significance of combining tutor talk moves and students' ITS performance were verified, as the extracted decision tree from the combined features significantly outperformed the trees resulting from the individual features.

Whereas it is intriguing to consider the possibility that these different types of talk (revoicing vs. reasoning) might benefit students with different levels of mastery on the ITS, our results are **correlational** and not causal. In particular, tutors may have perceived higher-knowledge students (i.e., those with higher ITS mastery) as more confident and capable of showing their reasoning, being asked about their math ideas, or seeing other students in their group provide explanations. Meanwhile, tutors may have perceived lower-knowledge students (i.e., those who needed more time or effort to master ITS content) as feeling more comfortable with the tutor being the one to build on students' expressed math ideas. Although the tutors centered discussions around student thinking in both cases, pressing for reasoning was predictive of achievement for students with higher content mastery, while revoicing predicted achievement for students with lower content mastery. These findings speak to the significance of future interventions in raising human educators' awareness about these two talk moves and their differential impact on students with different levels of content mastery.

**Limitations and Future Work** There are at least three caveats in our work. First, we did not analyze temporal dependencies between talk moves and ITS performance metrics. Future work should investigate possible causal and temporal relationships. Second, our approach of extracting decision trees from a random forest is not efficient as we prioritized more accurate models by exploring many candidate thresholds for splitting and by exhausting as many possible feature combinations without computational limitations. Future work should balance accuracy with efficiency. Finally, our analysis did not account for other student-specific factors, such as demographics, socio-economic status, and tutorial attendance, which are important directions for future work.

**Acknowledgments:** This research was supported by the Learning Engineering Virtual Institute (LEVI) program and the National Science Foundation (grants #2222647 and #1920510).